\documentclass[10pt,twocolumn,letterpaper]{article}

\usepackage{iccv}              

\definecolor{iccvblue}{rgb}{0.21,0.49,0.74}
\usepackage[pagebackref,breaklinks,colorlinks,allcolors=iccvblue]{hyperref}
\usepackage{graphicx}
\usepackage{amsmath}
\usepackage{multirow, float, booktabs, boldline, hhline}
\usepackage{algorithm}
\usepackage{listings}
\usepackage{etoolbox}
\makeatletter
\AfterEndEnvironment{algorithm}{\let\@algcomment\relax}
\AtEndEnvironment{algorithm}{\kern2pt\hrule\relax\vskip3pt\@algcomment}
\let\@algcomment\relax
\newcommand\algcomment[1]{\def\@algcomment{\footnotesize#1}}
\renewcommand\fs@ruled{\def\@fs@cfont{\bfseries}\let\@fs@capt\floatc@ruled
  \def\@fs@pre{\hrule height.8pt depth0pt \kern2pt}%
  \def\@fs@post{}%
  \def\@fs@mid{\kern2pt\hrule\kern2pt}%
  \let\@fs@iftopcapt\iftrue}
\makeatother

\usepackage{caption}
\usepackage{ctable}
\usepackage{diagbox}
\usepackage{cellspace}
\usepackage{color, colortbl}
\usepackage{multirow}
\usepackage{dsfont}
\usepackage{balance}
\usepackage{enumitem}

\usepackage{amsmath,amsfonts,bm}

\def\eqref#1{equation~\ref{#1}}

\def\1{\bm{1}}

\def\va{{\bm{a}}}

\def\ve{{\bm{e}}}

\def\vv{{\bm{v}}}
\def\vw{{\bm{w}}}
\def\vx{{\bm{x}}}

\def\vz{{\bm{z}}}

\def\mA{{\bm{A}}}

\def\mE{{\bm{E}}}

\def\mP{{\bm{P}}}

\def\mT{{\bm{T}}}

\def\mW{{\bm{W}}}

\def\mZ{{\bm{Z}}}

\DeclareMathAlphabet{\mathsfit}{\encodingdefault}{\sfdefault}{m}{sl}
\SetMathAlphabet{\mathsfit}{bold}{\encodingdefault}{\sfdefault}{bx}{n}

\usepackage{flushend}
\usepackage{balance}

\title{SVIP: Semantically Contextualized Visual Patches for Zero-Shot Learning}

\author{Zhi Chen$^{1}$\thanks{Equal contribution. Work done while at University of Queensland.}  \quad Zecheng Zhao$^{2*}$ \quad  Jingcai Guo$^3$ \quad  Jingjing Li$^4$ \quad Zi Huang$^2$ \\ $^1$University of Southern Queensland  $^2$University of Queensland \\ $^3$The Hong Kong Polytechnic University  \\  $^4$University of Electronic Science and Technology of China  \\ 
uqzhichen@gmail.com, helen.huang@uq.edu.au }

\begin{document}
\maketitle
\begin{abstract}
Zero-shot learning (ZSL) aims to recognize unseen classes without labeled training examples by leveraging class-level semantic descriptors such as attributes. A fundamental challenge in ZSL is semantic misalignment, where semantic-unrelated information involved in visual features introduce ambiguity to visual-semantic interaction. Unlike existing methods that suppress semantic-unrelated information post hoc either in the feature space or the model space, we propose addressing this issue at the input stage, preventing semantic-unrelated patches from propagating through the network. To this end, we introduce \textbf{\underline{S}}emantically contextualized \textbf{\underline{VI}}sual \textbf{\underline{P}}atches (\textbf{SVIP}) for ZSL, a transformer-based framework designed to enhance visual-semantic alignment. Specifically, we propose a self-supervised patch selection mechanism that preemptively learns to identify semantic-unrelated patches in the input space. This is trained with the supervision from aggregated attention scores across all transformer layers, which estimate each patch’s semantic score. As removing semantic-unrelated patches from the input sequence may disrupt object structure, we replace them with learnable patch embeddings. With initialization from word embeddings, we can ensure they remain semantically meaningful throughout feature extraction. Extensive experiments on ZSL benchmarks demonstrate that SVIP achieves state-of-the-art performance results while providing more interpretable and semantically rich feature representations. Code is available at \href{https://github.com/uqzhichen/SVIP}{https://github.com/uqzhichen/SVIP}.
\end{abstract}

\section{Introduction}
\label{sec:intro}
\begin{figure*}[t]
\centering
\includegraphics[width=1.9\columnwidth]{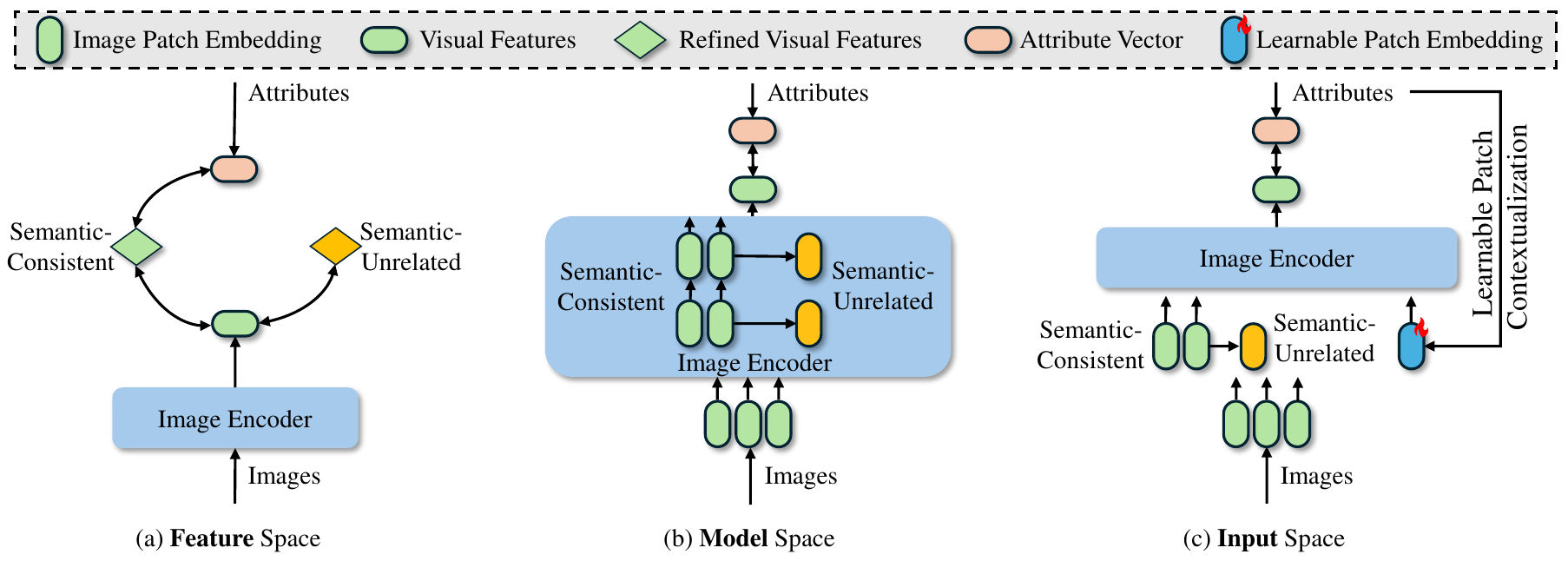}
\vspace{-5pt}
\caption{
Approaches to mitigate semantic misalignment for ZSL. (a) Disentangling methods remove semantic-unrelated information in the feature space. (b) Progressive semantic-guided methods remove the semantic-unrelated patches in the model space. (c) Our SVIP identifies the semantic-unrelated image patches in the input place and replace them with \textbf{semantically contextualized visual patches}.}
\label{FIG:intro}
\vspace{-10pt}
\end{figure*}

Zero-shot learning (ZSL) \cite{palatucci2009zero,xian2017zero,kodirov2017semantic} is a fundamental learning paradigm in computer vision that enables models to recognize classes unseen during training. ZSL methods typically recast classification as a regression task, where class-level semantic descriptors, such as attributes \cite{farhadi2009describing} serve as soft labels. For instance, each dimension of an attribute vector may indicate the presence or absence of a concept like the color pink. At test time, the predicted attribute vector is compared against the attributes of unseen classes, and the most similar class is selected.

Semantic misalignment is a fundamental challenge in ZSL \cite{tong2019hierarchical,han2020learning,chen2021free,chen2021semantics,chen2024progressive,li2021generalized}. Raw images often contain semantic-unrelated details that are not described by attributes. For example, background clutter, lighting variations, or surrounding objects can introduce misleading patterns. As a result, the model may latch onto unimportant features instead of the relevant semantic cues. Since these extraneous details are not explicitly annotated in attribute vectors, they disrupt the visual-semantic alignment by diluting key attributes such as color and shape. This makes it harder for the model to recognize unseen classes.

As shown in Figure \ref{FIG:intro}(a), prior work attempted to address this issue by refining or disentangling features, so that semantic-consistent information is preserved while semantic-unrelated information is suppressed \cite{chen2021semantics,chen2021free,han2020learning,li2021generalized}. Thus, the preserved semantic-consistent visual features can enhance visual-semantic alignment. Although these post hoc refinements help, they cannot fully prevent extraneous information from seeping into the learned representations.

In parallel, transformer-based architectures \cite{waswani2017attention} have emerged as powerful backbones in vision tasks \cite{han2022survey} by modeling images as sequences of patches. In light of the strong representation learning capability of Vision Transformers (ViTs) \cite{dosovitskiy2020image,yuan2021tokens}, researchers in ZSL community are turning into designing ViT-based methods \cite{chen2024progressive,cheng2023hybrid,liu2023progressive}. The patch modelling behavior enables identifying the semantic-unrelated patches. As in Figure \ref{FIG:intro}(b), ZSLViT \cite{chen2024progressive} progressively prunes visual tokens that do not map to semantics in the model space. However, research has shown that semantic features can become diluted in deeper layers of ViTs \cite{qin2022devil}, making it easier for semantic-unrelated information to persist and potentially contaminate the final visual features.

As above, these methods either remove the semantic-unrelated features after extracting the entangled visual features, or during extracting the visual features, they do not entirely prevent such features from influencing learned representations. Consequently, an intriguing research question arises: \textbf{Can we preemptively remove or repurpose semantic-unrelated patches at the input stage to avoid entangling them in the learned features? }

To achieve this goal, we propose \textbf{SVIP}, \textbf{\underline{S}}emantically contextualized \textbf{\underline{VI}}sual \textbf{\underline{P}}atch, a ViT-based framework tailored for mitigating the semantic-unrelated information and enhance semantic-visual propagation in ZSL. As shown in Figure \ref{FIG:intro}(c), unlike prior work that prunes or refines features after they have been extracted, SVIP identifies and contextualizes these patches upfront.

Specifically, to mitigate the semantic misalignment problem comprehensively, we propose self-supervised patch selection mechanism that learns to identify the semantic-unrelated patches in the input space, \textit{i.e.,} before feeding into the transformer blocks. In particular, we aggregate the self-attention weight matrices across all transformer blocks, yielding reliable semantic understanding about the patch importance throughout the entire ViT. To mitigate the negative impact of semantic-unrelated patches to semantic-related patches, we train a patch classifier to determine the input patches that are semantic-unrelated patches, using the supervision from the aggregated attention matrix. At the test time, the patch classifier can directly prevent the semantic-unrelated patches from propagating into the feature extraction. Once the semantic-unrelated patches are identified in the input sequence, a naive solution is removing them from the input sequence. However, this can break object structure on the image. Instead, we propose to mask the semantic-unrelated patches with a learnable semantic context patch. The semantic context patch is initialized with the attribute-level word embeddings that provides rich semantic information. These contextualize input patches can enhance the visual-semantic interaction throughout the visual feature extraction process. Overall, the contributions of this paper are:

\begin{itemize}
\item We introduce \textbf{\underline{S}}emantically contextualized \textbf{\underline{VI}}sual \textbf{\underline{P}}atch (\textbf{SVIP}), a ViT-based framework for addressing the semantic misalignment problem in zero-shot learning. 
It emphasizes the early-stage handling of semantic-unrelated patches to achieve a more focused and unambiguous feature space for ZSL.
\item We propose a self-supervised patch selection strategy to identify semantic-unrelated patches using the attention scores aggregated from the self-attention weight matrix across the ViT blocks.  We propose patch semantic contextualization that injects attribute-level word embeddings into semantic-unrelated patches, ensuring that semantic information is incorporated into visual representations from the initial layers of the transformer.
\item  SVIP achieves state-of-the-art performance results on three ZSL benchmark datasets on both ZSL and generalized ZSL settings, demonstrating the effectiveness in mitigating semantic misalignment in the input space.
\end{itemize}

\section{Related Work}
\label{sec:relatedwork}
In the deep learning era \cite{you2021domain,chen2021local,wang2023cal,you2022pixel,wei2024plantseg,wei2024benchmarking,wang2025discrimination,limdipex,zhang2024towards,chen2024fastedit,chen2024cf,yu2025dynamic,zhao2025continual,zhang2025tokenbinder,zhao2025synthetic,zhang2025umivr,wang2024search,wang2022discovering,wang2024tokens}, the core objective in ZSL  \cite{romera2015embarrassingly,xian2017zero,kodirov2017semantic,xian2018feature,chen2021semantics,chen2020rethinking,su2022distinguishing,chen2021entropy,chen2023zero,guo2024fine,guo2024element,chen2022federated,guo2020novel,guo2023graph} is to generalize visual-semantic knowledge obtained from seen classes to unseen classes. This generalization relies on semantic descriptors, which are typically represented as attributes~\cite{farhadi2009describing}, documents~\cite{elhoseiny2013write}, or word embeddings~\cite{norouzi2013zero}. The mainstream ZSL methods can be generally categorized as generative methods and embedding-based methods.
Generative ZSL methods typically synthesize unseen visual features from semantic descriptors using generative models, enabling standard classification pipelines to incorporate synthetic training samples~\cite{xian2018feature,yu2020episode,chen2021semantics}. These approaches leverage various generative architectures, including GANs \cite{xian2018feature,zhu2018generative,su2022distinguishing}, VAEs \cite{wang2018zero,schonfeld2019generalized} and Normalizing Flows \cite{shen2020invertible,chengsmflow,chen2021mitigating}. 
Embedding-based ZSL methods aim to learn a direct mapping from the visual to the semantic space, aligning image representations with class descriptors. These methods typically leverage the attention mechanism to localize the semantic descriptors. Zhu \textit{et al.} \cite{zhu2019semantic} introduced a semantic-guided multi-attention localization network to extract the local fine-grained features of the discriminative patches. Xu \textit{et al.} \cite{xu2020attribute} proposed an attribute prototype network, where each prototype represents a specific semantic concept and is used as the 1×1 filter to convolve the feature maps with the feature maps from ResNet101 in order to attend to the region of interest. The follow-up work \cite{liu2021goal} proposed to use the GloVe \cite{pennington2014glove} model to extract semantic vectors from the attribute names (\textit{e.g.,} ``white breast"), which are further used to represent the attribute prototypes. More recently, Chen \textit{et al.} \cite{chen2022msdn} developed a mutually semantic distillation network to refine the alignment between visual and semantic features.

Despite advancements in generative and embeddings-based methods, semantic misalignment remains a fundamental challenge in ZSL \cite{han2020learning,chen2021free,chen2021semantics,chen2024progressive,li2021generalized,chen2020canzsl}. Several works attempt to disentangle semantic-related from semantic-unrelated features. DLFZRL \cite{tong2019hierarchical} was the first to introduce disentangling-based ZSL, splitting image representations into semantic, non-semantic, non-discriminative features. RFF \cite{han2020learning} applied mutual information minimization to remove redundant visual details, while SDGZSL \cite{chen2021semantics} utilized total correlation analysis to separate semantic-consistent and semantic-unrelated components in the feature space. FREE \cite{chen2021free} refined visual features through a self-adaptive margin center loss to enhance semantic alignment. Dis-VAE \cite{li2021generalized} introduced a batch recombination strategy to distill category-specific features for more accurate recognition. Recent Vit-based method ZSLViT \cite{chen2024progressive} introduces a vision transformer-based ZSL framework that merge those semantic-unrelated tokens to improve the semantic-visual alignment in learning ZSL representations. Despite their efforts in mitigating the semantic-alignment problem in either feature or model spaces, the power of those trial visual patches that are unattended in attention mechanism remains neglected. In this paper, we learn to identify the semantic-unrelated visual patches at the input space and semantically contextualize these patches, thus allowing them to propagate the semantic information onto those important patches throughout the visual feature learning stages. 

\begin{figure*}[t]
    \centering
    \includegraphics[width=1.0\linewidth]{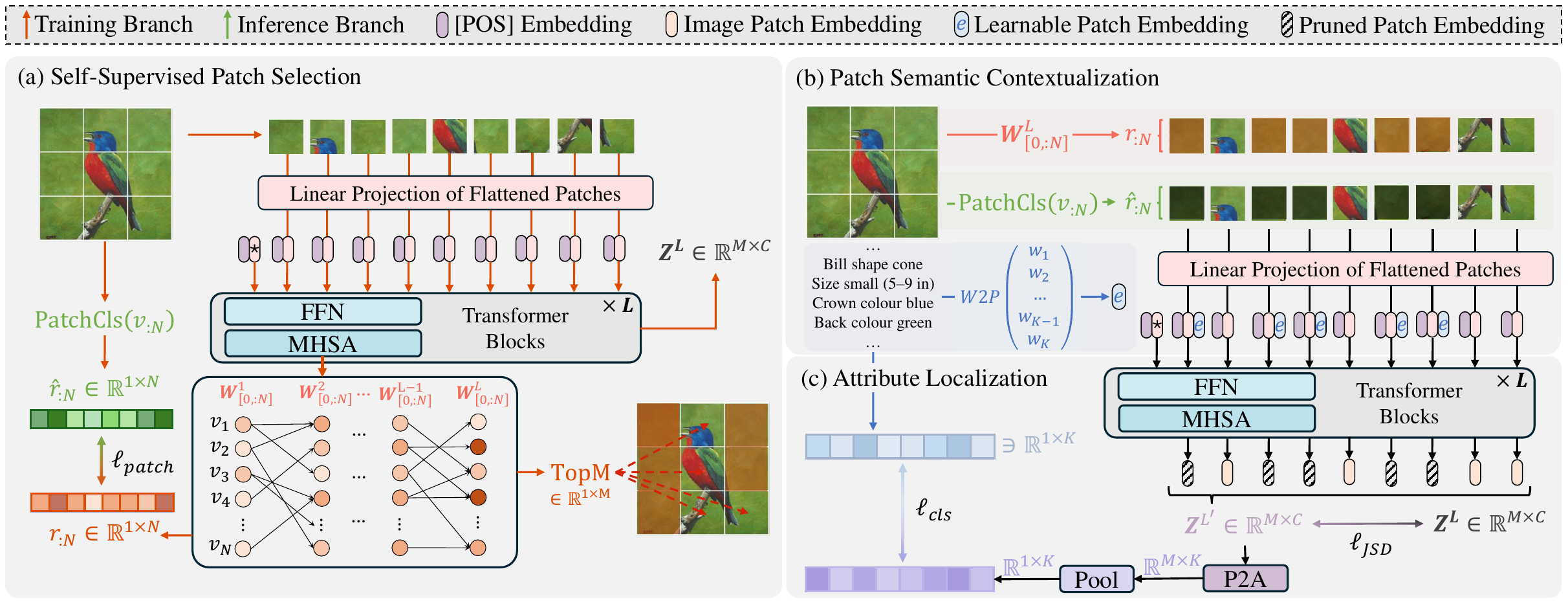}
    \caption{The framework of our proposed SVIP, which includes three components: (a) Self-Supervised Patch Selection: we aggregate attention matrices throughout transformer blocks to guide the training of patch classifier in identifying the semantic-unrelated input patches. (b) Patch Semantic Contextualization: we apply learnable patch embeddings initialized with attribute word embeddings to contextualize the semantic-unrelated patches. (c) Attribute Localization: we localize the attribute values with final output patch embeddings, instead of the class token. }
    \label{fig:Model_Architecture}
    \vspace{-10pt}
\end{figure*}

\begin{figure}[t]
	\centering
	\includegraphics[width=1\columnwidth]{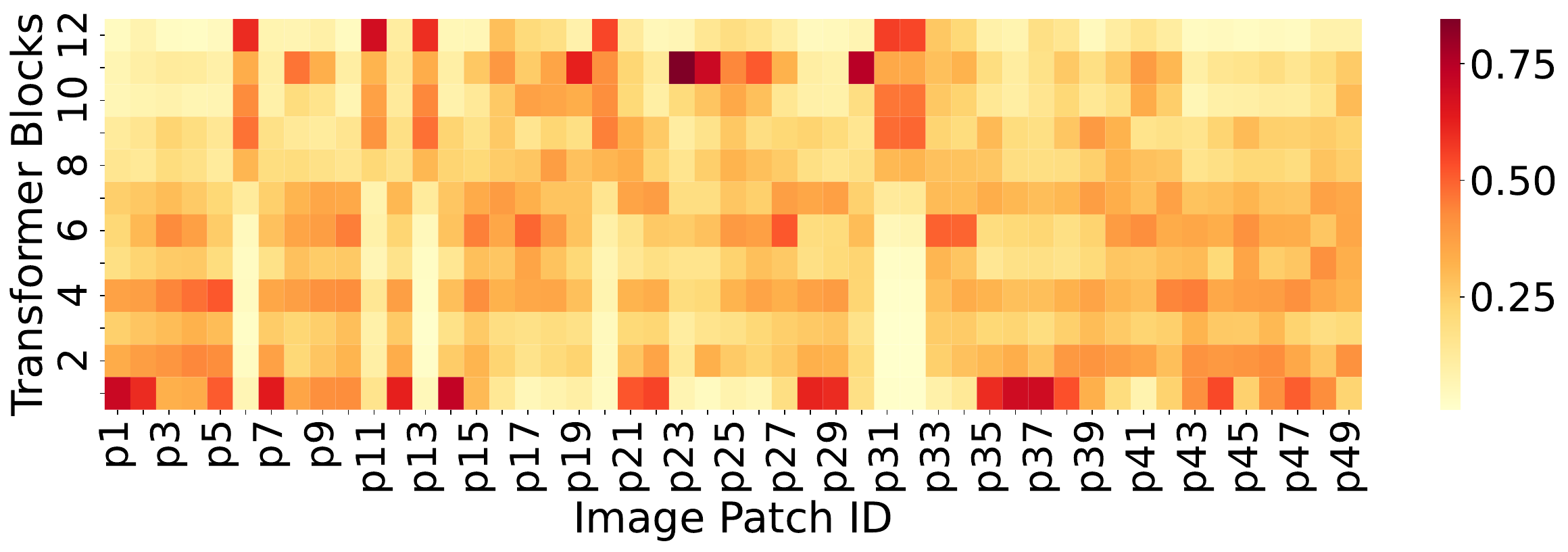}
        \vspace{-20pt}
	\caption{A visualization of patch attention weights across all self-attention layers in ViT.}
	\label{FIG:att_map}
    \vspace{-10pt}
\end{figure}

\section{Methodology}
\subsection{Problem Formulation}
Let $D^s = \{(\vx^s_i, y^s_i)\}^{n_s}$ be a \textbf{seen} dataset with $C^s$ classes,  where each image $\vx^s_i \in \mathcal{X}$ has a corresponding label $y_i^s \in \mathcal{Y}$. We also have an \textbf{unseen} dataset $D^u= \{(\vx^u_i, y^u_i)\}^{n_u}$ with $C^u$ classes. By design, $C^s$ and $C^u$ are disjoint, ensuring $C^s \cup C^u = \emptyset$. We denote the semantic information, \textit{i.e.,} attributes by $\mathcal{A} = \{\va_y\}^{|C^s+C^u|}_{y=1}$ for all classes, while only the seen classes are available for training. Our goal is to train a model on $D^s$ such that it can accurately recognize classes in $C^u$ at test time, without ever seeing labeled examples for $C^u$ during training.

\subsection{Motivation}
ZSL hinges on aligning visual features with semantic information. However, semantic-unrelated information in the visual space complicates this alignment. It creates ambiguity in how attributes map to relevant image regions. As shown in Figure~\ref{FIG:intro}, existing methods remove semantic-unrelated information in either (1) the \textbf{feature space}, by post-processing extracted features, or (2) the \textbf{model space}, by pruning irrelevant patches throughout the different layers. While these strategies partially suppress semantic-unrelated information, they fail to eliminate them altogether. Thus, such noise persists in the final representations.

In contrast, as shown in Figure \ref{fig:Model_Architecture}, we propose identifying and filtering semantic-unrelated patches in the \textbf{input space}, before it propagates into the model. This early intervention preserves a cleaner, more semantically coherent representation throughout feature extraction. Consequently, the subsequent ZSL pipeline can concentrate on meaningful visual-semantic interactions and improve unseen class recognition.

\subsection{Self-Supervised Patch Selection}
To localize the semantic-unrelated patches, we propose a Self-Supervised Patch Selection (SSPS) mechanism that trains a binary classifier to predict the semantic score of image patches, using the aggregated attention from the transformer as supervision. 

\vspace{3pt}
\noindent \textbf{Revisiting ViT.}
We partition an image $\vx \in \mathbb{R}^{H\times W\times C}$ into $N$ patches $\{\mP_1,\cdots,\mP_N\}$. Each patch is then flattened and linearly projected as a patch embedding:
\begin{equation}
\begin{aligned}
\{\vv_1,\cdots,\vv_N\} = \texttt{PatchEmb}(\{\mP_1,\cdots,\mP_N\}),
\label{eq:1}
\end{aligned}
\end{equation}
we prepend a learnable class token $\vv_{\text{cls}}$ and add positional embeddings $\mE_{pos}$ to form the initial patch sequence:
\begin{equation}
\begin{aligned}
\mZ^0 = [\vv_{\text{cls}},\vv_1,\cdots,\vv_N] + \mE_{pos}.
\label{eq:2}
\end{aligned}
\end{equation}
This patch sequence passes through $L$ transformer blocks:
\begin{equation}
\begin{aligned}
\mZ^l = \text{TransformerBlock}^l(\mZ^{l-1}), ~~~ l=1,\cdots,L.
\label{eq:3}
\end{aligned}
\end{equation}
yielding final outputs $\mZ^L=[\vz_{\text{cls}}, \vz_1,\cdots,\vz_N]$. 
Here, the class token $\vz_{\text{cls}}$ is taken as the global representation.

\vspace{5pt}
\noindent \textbf{Patch Supervision via Attention Matrix Aggregation.}
To estimate each patch's semantic score, we gather attention scores from the multi-head self-attention (MHSA) blocks. Let $\mT^l \in \mathbb{R}^{(N+1)\times(N+1)}$ be the attention matrix in the $l$-th transformer block summed over all attention heads. We aggregate these attention scores across layers:
\begin{equation}
\begin{aligned}
\mW^l = \mW^{l-1} + \mW^{l-1}\times \mT^{l}, ~~~ l=1,\cdots,L,
\label{eq:4}
\end{aligned}
\end{equation}
where $\mW^1 = \mT^1$. The aggregated matrix $\mW^L$ offers a global view of patch importance. We further take the row corresponding to the class token ($0$-th index) to provide each patch's score: $r_i = \mW^L_{[0;i]}$, which can serve as pseudo ground truth of semantic scores.

We then introduce an auxiliary patch classifier to predict the semantic score of each patch embedding: 
\begin{equation}
\begin{aligned}
\hat{r}_i =\texttt{PatchCls($\vv_i$)}, ~~~ i = 1,\cdots,N.
\label{eq:5}
\end{aligned}
\end{equation}
Aligning $\hat{r}_i$ with the attention-derived $r_i$ allows the model to learn fine-grained, semantic patch discrimination with a binary cross entropy loss:
\begin{equation}
\begin{aligned}
\mathcal{L}_{\text{patch}} = -\frac{1}{N} \sum_{i=1}^{N} \Big[ r_i \log \hat{r}_i + (1 - r_i) \log (1 - \hat{r}_i) \Big].
\label{eq:6}
\end{aligned}
\end{equation}

This design is motivated by the dynamic nature of attention scores across the transformer layers. As shown in Figure \ref{FIG:att_map}, we visualize the attention score of the 49 patches across 12 transformer layers. The importance of the these patches are not consistent across different layers. Thus, relying on certain layers to determine the semantic-unrelated patches may lead to misclassification. In SSPS, we produce more robust pseudo labels by aggregating the attention scores across all layers.

\subsection{Patch Semantic Contextualization}
While removing semantic-unrelated patches might preserve only informative patches, it can disrupt object structures. Instead, we propose Patch Semantic Contextualization (PSC) to contextualize these patches with attribute-level information, transforming their embeddings to better align with semantics.

\noindent \textbf{Learnable Patch Embedding.} 
We define a learnable patch embedding $\ve$ initialized from word vectors \cite{pennington2014glove}. Let $\{\vw_1,\cdots,\vw_K\}$ be the word embeddings for $K$ attributes (\textit{e.g.,} color, shape). These are aggregated via a Word-to-Patch (\texttt{W2P}) projection layer:
\begin{equation}
\begin{aligned}
\ve = \texttt{W2P}(\vw_1,\cdots,\vw_K).
\label{eq:7}
\end{aligned}
\end{equation}
Next, we identify patches whose predicted scores $\hat{r}_i$ exceed a threshold or rank within the top $M$:
\begin{equation}
\begin{aligned}
\mathcal{S}_M = \operatorname{TopM}_{i \in \{1,\dots,N\}} \{\hat{r}_i, M\}.
\label{eq:8}
\end{aligned}
\end{equation}
We then inject $\ve$ into each selected patches like injecting positional embeddings:
\begin{equation}
\begin{aligned}
\mZ^{0'} = [\vv_{\text{cls}},\vv_1^{'},\cdots,\vv_N^{'}] + \mE_{pos}, \\
\hat{\vv}_i =
\begin{cases}
\vv_i, & \text{if } i \in \mathcal{S}_M \\
\vv_i + \ve, & \text{otherwise}.
\end{cases}
\label{eq:9}
\end{aligned}
\end{equation}
Because these updated patch embeddings $\hat{\vv}_i$ carry attribute-level semantic information in the input space, which can enhance the visual-semantic interaction throughout transformer blocks.

\begin{algorithm}[t]
\caption{Pseudocode of SVIP in a PyTorch-like style.}
\label{alg:code}
\footnotesize\textit{\textbf{Input:} imgs – batch of input images, A – attribute matrix for seen classes, WordEmb – word embeddings for attribute labels, labels – GT labels}
\definecolor{codeblue}{RGB}{96,125,139}
\definecolor{sec}{RGB}{210,130,70}
\lstset{
  backgroundcolor=\color{white},
  basicstyle=\fontsize{7.2pt}{7.2pt}\ttfamily,
  columns=fullflexible,
  breaklines=true,
  captionpos=b,
  commentstyle=\fontsize{7.2pt}{7.2pt}\color{codeblue},
  keywordstyle=\fontsize{7.2pt}{7.2pt},
  escapeinside={(*@}{@*)}, 
}
\begin{lstlisting}[language=python]
for imgs in loader: 
    (*@\textcolor{sec}{ --- Self-Supervised Patch Selection (Sec 3.3) ---}@*)
    # project images to patch embeddings: BxNxC
    pt_emb = PatchEmbed(imgs) 
    # patch importance logits
    pt_logits = PatchCls(pt_emb)
    # attention maps & transformer features
    att_map, pt_out = Transformers.forward(pt_emb)
    # aggregate attn to get patch labels
    pt_label = attention_agg(att_map)
    # self-supervised patch loss, Eq.(6)
    loss = BCELoss(pt_logits, pt_label)
    
    (*@\textcolor{sec}{ --- Patch Semantic Contextualization (Sec 3.4) ---}@*)
    # project to attribute dims & pooling 
    attr_emb = pooling(P2A(pt_emb[pt_label]))
    # classification logits via matrix multiplication 
    cl_logits1 = mm(att.view(B,C), A.view(C,K))
    # cross entropy loss, Eq.(12)
    loss += CrossEntropyLoss(cl_logits1/t, labels)
    # project word to patch embeddings
    e_emb = W2P(WordEmb)
    # contextualize semantic-unrelated patches
    pt_emb[pt_label] += e_emb
    # update patch embeddings: output patches BxMxC
    _, pt_emb = Transformers.forward(pt_emb) 
    
    (*@\textcolor{sec}{ --- Attribute Localization (Sec 3.5) ---}@*)
    # project to attribute dims & pooling 
    attr_emb = pooling(P2A(pt_emb[pt_label]))
    # classification logits
    cl_logits2 = mm(attr_emb.view(B,C), A.view(C,K))
    # cross entropy loss, Eq.(12)
    loss += CrossEntropyLoss(cl_logits2/2, labels)
    # JS Divergence between predictions, Eq.(13)
    loss += JSDivergence(cl_logits1, cl_logits2)
    
    # SGD update: backpropagation & update
    loss.backward()
    update(PatchEmbed.params, Transformers.params, PatchCls.params, W2P.params, P2A.params)
\end{lstlisting}
\vspace{-3pt}
\end{algorithm}

\subsection{Attribute Localization}
Although the final class token $\vv_{cls}$ can capture the global representation, the semantic-related patches can help localize attributes.  Let $\mZ^{L'} = [\vz_i^{'} | i \in \mathcal{S}_M]$ be the final representations of the top $M$ semantic-related patches. We learn a Patch-to-Attribute projection function (\texttt{P2A}) to map the patch embeddings $ \mZ^{L'} \in \mathbb{R}^{M\times C}$ into the attribute space: $\hat{\mA} = \texttt{P2A}(\mZ^{L'})\in \mathbb{R}^{M\times K}$, where $K$ is the number of attributes that describes a category. 
A max pooling operation is then applied over these patch-attribute embeddings: 
\begin{equation}
\begin{aligned}
\hat{\va} = MaxPool(\hat{\mA}),
\label{eq:10}
\end{aligned}
\end{equation}
which identifies the patch most relevant for each attribute. Thus, the model learns to focus when predicting an attribute value. The classification probability follows a standard SoftMax over cosine similarities:
\begin{equation}
\begin{aligned}
p(y|\mZ^{0}) = \frac{\exp(\sigma~\text{cos}(\hat{\va}, \va^s_y))}{\sum_{\hat{y}}\in \mathcal{Y}^s \exp (\sigma~\text{cos}(\hat{\va},  \va^s_{\hat{y}}))},
\label{eq:11}
\end{aligned}
\end{equation}
where $\va^s_y$ is the attribute vector of a seen class $y$ and $\sigma$ is the temperature for scaling the cosine similarities.

\subsection{Model Optimization and Inference}
\noindent \textbf{Optimization.} 
We propagate each sample through the transformer blocks for twice: once with the original patches $\mZ^{0}$ and once with the contextualized patches $\mZ^{0'}$. Hence, out classification loss sums both predictions:
\begin{equation}
\begin{aligned}
\ell_{cls} = - \log p(y|\mZ^{0}) - \log p(y|\mZ^{0'}).
\label{eq:12}
\end{aligned}
\end{equation}
To stabilize the training for contextualized patches, we apply the Jensen–Shannon divergence between the two output distributions:
\begin{equation}
\begin{aligned}
\ell_{JSD} = \frac{1}{2} \sum_{i} p(y|\mZ^{0}) \log \frac{p(y|\mZ^{0})}{p(y|\mZ^{0'})} \\+~ \frac{1}{2} \sum_{i} p(y|\mZ^{0'}) \log \frac{p(y|\mZ^{0'})}{p(y|\mZ^{0})}.
\label{eq:13}
\end{aligned}
\end{equation}
Overall, the training objective is defined as:
\begin{equation}
\begin{aligned}
\ell_{overall} = \ell_{cls} + \lambda_1 \ell_{JSD} + \lambda_2 \ell_{patch},
\label{eq:14}
\end{aligned}
\end{equation}
where $\lambda_1$ and $\lambda_2$ are the loss coefficients.
The overall training process is illustrated in Algorithm \ref{alg:code}. 

\noindent \textbf{Zero-Shot Inference.} 
We first partition a test image into multiple patches and project them into corresponding patch embeddings using Eq. \ref{eq:1}. Then, we identify the semantic-unrelated patches with the learned patch classifier using Eq. \ref{eq:5}. While all patch embedding are applied with positional embeddings, only those semantic-unrelated patches are applied with learned semantic patch embeddings as in Eq. \ref{eq:9}. These patch embeddings are fed into transformer blocks and for further attribute localization. Lastly, we infer the predicted unseen class label using Eq. \ref{eq:11}.

\label{sec:method}

\begin {table*}[t]
\caption {
Comparison of state-of-the-art ZSL and GZSL models on three benchmark datasets. For ZSL, results are reported as average top-1 accuracy (T1). For GZSL, we report top-1 accuracy for unseen (U) and seen (S) classes, along with their harmonic mean (H). The best and second-best results are highlighted in \textbf{bold} and \underline{underlined}, respectively.
}

\centering
\scalebox{0.98}{
\begin{tabular}[t]{@{~}cl@{~}c@{~}|cccc|cccc|cccc}
\hline
  \multirow{2}{*}{}     & & & \multicolumn{4}{c|}{\textbf{CUB}}  &  \multicolumn{4}{c|}{\textbf{AwA2}}   & \multicolumn{4}{c}{\textbf{SUN}}\\ 
  
\hhline{|~~~|------------|}
& \multirow{-2}{*}{\textbf{Methods}} & \multirow{-2}{*}{\textbf{Venue}} & \multicolumn{1}{c|}{\textit{T1}} &\textit{U} & \textit{S} & \textit{H}  & \multicolumn{1}{c|}{\textit{T1}} & \textit{U} & \textit{S} & \textit{H}  & \multicolumn{1}{c|}{\textit{T1}} & \textit{U} & \textit{S} &        \textit{H} \\
    
\hline
\multirow{8}{*}{\rotatebox{270}{ResNet101}} 

&RFF  \cite{han2020learning}&   CVPR'20     
&\multicolumn{1}{c|}{-}    & 52.6          & 56.6    & 54.6
&\multicolumn{1}{c|}{-}    & 59.8          & 75.1    & 66.5   
&\multicolumn{1}{c|}{-}    & 45.7          & 38.6    & 41.9\\ 

& CANZSL \cite{chen2020canzsl}  &WACV'20 
&\multicolumn{1}{c|}{60.6}    & 47.9              & 58.1              & 52.5
&\multicolumn{1}{c|}{68.9}    & 49.7              & 70.2              & 58.2
&\multicolumn{1}{c|}{60.1}    & 46.8              & 35.0              & 40.0
\\

&SDGZSL \cite{chen2021semantics}    &  ICCV'21   
&\multicolumn{1}{c|}{75.5}    & 59.9        & 66.4      & 63.0
&\multicolumn{1}{c|}{72.1}    & 64.6        & 73.6      & 68.8    
&\multicolumn{1}{c|}{62.4}    & 48.2        & 36.1      & 41.3\\

&Dis-VAE \cite{li2021generalized}    &  AAAI'21   
&\multicolumn{1}{c|}{-}    & 51.1        & 58.2      & 54.4
&\multicolumn{1}{c|}{-}    & 56.9        & 80.2      & 66.6    
&\multicolumn{1}{c|}{-}    & 36.6        & 47.6      & 41.4\\

&
HSVA \cite{chen2021hsva}      &NeurIPS'21
&\multicolumn{1}{c|}{62.8}    & 52.7        & 58.3      & 55.3
&\multicolumn{1}{c|}{-}       & 59.3        & 76.6      & 66.8
&\multicolumn{1}{c|}{63.8}    & 48.6        & 39.0      & 43.3 \\
&
MSDN  \cite{chen2022msdn}&      CVPR'22   
&\multicolumn{1}{c|}{76.1} & 68.7          & 67.5     & 68.1
&\multicolumn{1}{c|}{70.1} & 62.0          & 74.5     & 67.7    
&\multicolumn{1}{c|}{65.8} & \underline{52.2} & 34.2 & 41.3 
\\

& 
ICIS \cite{christensen2023image}  & ICCV’23 
& \multicolumn{1}{c|}{60.6} & 45.8 & 73.7 & 56.5 
& \multicolumn{1}{c|}{64.6} & 35.6 & \textbf{93.3} & 51.6
& \multicolumn{1}{c|}{51.8} & 45.2 & 25.6 & 32.7 \\ 

& HAS \cite{chen2023zero} & MM'23
&\multicolumn{1}{c|}{76.5} & 69.6 & 74.1 & 71.8   
&\multicolumn{1}{c|}{71.4} & 63.1 & 87.3 & 73.3
&\multicolumn{1}{c|}{63.2} & 42.8 & 38.9 & 40.8
\\

    \hline

\multirow{7}{*}{\rotatebox{270}{ViT-based}}

&CLIP  \cite{radford2021learning}&      ICML'21      
&\multicolumn{1}{c|}{-} & 55.2          & 54.8     & 55.0
&\multicolumn{1}{c|}{-} & -             & -        & -   
&\multicolumn{1}{c|}{-} & -             & -        & - 
\\

&CoOp  \cite{zhou2022learning}&      IJCV'22     
&\multicolumn{1}{c|}{-} & 49.2          & 63.8     & 55.6
&\multicolumn{1}{c|}{-} & -             & -        & -   
&\multicolumn{1}{c|}{-} & -             & -        & - 
\\

&I2DFormer  \cite{naeem2022i2dformer}&      NeurIPS'22
&\multicolumn{1}{c|}{45.4} & 35.3          & 57.6     & 43.8
&\multicolumn{1}{c|}{\textbf{76.4}} & \textbf{66.8}          & 76.8     & 71.5   
&\multicolumn{1}{c|}{-} & -             & -        & - 
\\

&I2MV  \cite{naeem2023i2mvformer}&      CVPR'23    
&\multicolumn{1}{c|}{42.1} & 32.4          & 63.1     & 42.8
&\multicolumn{1}{c|}{\underline{73.6}} & \underline{66.6}          & 82.9     & 73.8    
&\multicolumn{1}{c|}{-}    & -             & -        & - 
\\
&
DUET  \cite{chen2023duet} &      AAAI'23      
&\multicolumn{1}{c|}{72.3} & 62.9          & 72.8     & 67.5
&\multicolumn{1}{c|}{69.9} & 63.7          & 84.7     & 72.7    
&\multicolumn{1}{c|}{64.4} & 45.7          & 45.8        & 45.8 
\\

&ZSLViT  \cite{chen2024progressive}&     CVPR'24  
&\multicolumn{1}{c|}{\underline{78.9}}    
& \underline{69.4}          
& \textbf{78.2}              
& \underline{73.6}
&\multicolumn{1}{c|}{70.7} 
& {66.1}          
& 84.6      
& \underline{74.2}
&\multicolumn{1}{c|}{\underline{68.3}}   
& 45.9          
& \textbf{48.4}          
& \underline{47.3} 
\\

\hhline{|~|--------------|}
&
\textbf{SVIP}(ours) & ICCV'25
& \multicolumn{1}{c|}{\textbf{79.8}} 
& \textbf{72.1} 
& \underline{78.1} 
& \textbf{75.0}   
& \multicolumn{1}{c|}{{69.8}} 
& 65.4
& \underline{87.7} 
& \textbf{74.9}
&\multicolumn{1}{c|}{\textbf{71.6}} 
& \textbf{53.7} 
& \underline{48.0} 
& \textbf{50.7}
\\

\hline
\end{tabular}}
\label{gzslperoformance}
\vspace{-5pt}
\end {table*}

\section{Experiments}

\subsection{Experimental Setting}

\noindent \textbf{Datasets.}
We conduct experiments on three zero-shot learning benchmark datasets, \textit{i.e.,} CUB \cite{wah2011caltech}, AwA2 \cite{lampert2013attribute}, SUN \cite{patterson2014sun}.
Specifically, CUB contains 11,788 images over 200 bird categories, each annotated with 312 attributes. AwA2 comprises 37,322 images across 50 animal species described by 85 attributes. SUN consists of 14,340 images from 717 scene classes captured with 102 attributes. We use the training splits proposed in \cite{xian2018feature}, the seen/unseen classes are split into 150/50, 40/10 and 645/72 classes respectively.

\vspace{5pt}
\noindent \textbf{Evaluation.} Following \cite{xian2018zero}, we measure the top-1 accuracy in both zero-shot learning (ZSL) and generalized zero-shot learning (GZSL) settings. In the ZSL setting, we only evaluate the accuracy of the test samples in the unseen classes. In the GZSL setting, we compute the accuracy of the test samples from both the seen (denoted as $S$) and unseen classes (denoted as $U$). To generally evaluate the performance, the harmonic mean between seen and unseen classes is a main evaluation metric in the GZSL setting, defined as $H = (2\times S\times U)/(S + U)$.

\vspace{5pt}
\noindent \textbf{Implementation.}
Our model is implemented based on the ViT-base model \cite{touvron2021training} (image input size: 224$\times$224), pretrained on ImageNet-1k as a baseline and for initialization, which is consistent with existing ViT-based ZSL methods. ViT-base has a default patch number of 196, but for computational efficiency, we aggregate each 4 (2$\times$2) patches as one, resulting in $49$ patches. 

\begin{table*}[]
\centering
\caption{Analysis of different components in SVIP. }
\label{Tab:ablations}
\begin{tabular}{ @{~~~~} l @{~~~~} | @{~~~~} c c c c @{~~~~} | @{~~~~} c c c c @{~~~~} | @{~~~~} c c c c }
    \hline
    \multirow{2}{*}{Method}  
    & \multicolumn{4}{c}{CUB} 
    & \multicolumn{4}{c}{AwA2}
    & \multicolumn{4}{c}{SUN} \\ 
    
    \hhline{~------------}
     & $T1$ & $U$ & $S$ & $H$ & $T1$ & $U$ & $S$ & $H$ & $T1$ & $U$ & $S$ & $H$ \\ \hline \hline
    
    Baseline(ViT w/ att head)   
    & 76.8  & 59.8  & 68.4  & 63.8    
    & 61.4  & 58.0  & 81.6  & 67.8  
    & 62.7  & 43.8  & 30.6  & 36.0  \\
                    
    SVIP w/o SSPS (Eq. \ref{eq:6})  & 78.9  & 68.8  & 75.3  & 71.9  
                                    & 66.8  & 64.7  & 82.6  & 72.6  
                                    & 67.6  & 48.8  & 45.9  & 47.3\\
    
    SVIP w/o PSC (Eq. \ref{eq:9})   & 78.1  & 69.3  & 76.2  & 72.6  
                                    & 67.6  & 64.2  & 83.1  & 72.4  
                                    & 67.9  & 49.2  & 46.8  & 48.0 \\
    
    SVIP w/o JSD (Eq. \ref{eq:13})  & 79.5  & 71.9  & 78.1  & 74.9  
                                    & 69.1  & 65.3  & 86.7  & 74.5 
                                    & 71.2  & 53.2  & 47.9  & 50.4\\
    
    SVIP w/o W2P (Eq. \ref{eq:7})   & 79.1  & 70.8  & \textbf{78.6}  & 74.5  
                                    & 69.8  & 64.9  & 87.3  & 74.4 
                                    & 71.5  & 53.1  & 47.5  & 50.1\\   
    
    SVIP w/o P2A  (Eq. \ref{eq:10}) & 78.5  & 69.4  & 76.8  & 72.9  
                                    & 69.5  & 65.0  & \textbf{88.1}  & 74.8 
                                    & 70.1  & 53.8  & 47.5  & 50.5\\
    
    SVIP (full)                     & \textbf{79.8}  & \textbf{72.1}  & {78.1}  &        \textbf{75.0}  
                                    & \textbf{69.8}  & \textbf{65.4}  & {87.7}  & \textbf{74.9}   
                                    & \textbf{71.6}  & \textbf{53.7}  & \textbf{48.0}  & \textbf{50.7}   \\ \hline
\end{tabular}
\end{table*}

\begin{figure*}[t]
    \centering
    \includegraphics[width=1.0\linewidth]{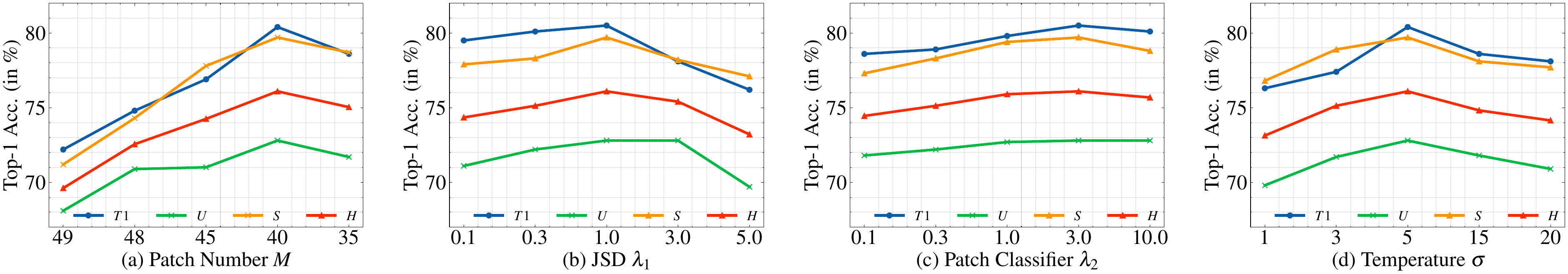}
    \vspace{-10pt}
    \caption{Hyper-parameter sensitivity on CUB dataset. The horizontal axis indicates the varying hyper-parameters for (1) the preserved patch number M, (2) JSD loss coefficient $\lambda_1$, (3) patch classifier loss coefficient $\lambda_2$, and (4) logits temperature $\sigma$.}
    \label{hyper}
    \vspace{-5pt}
\end{figure*}

\begin{figure*}[t]
    \centering
    \includegraphics[width=1.0\linewidth]{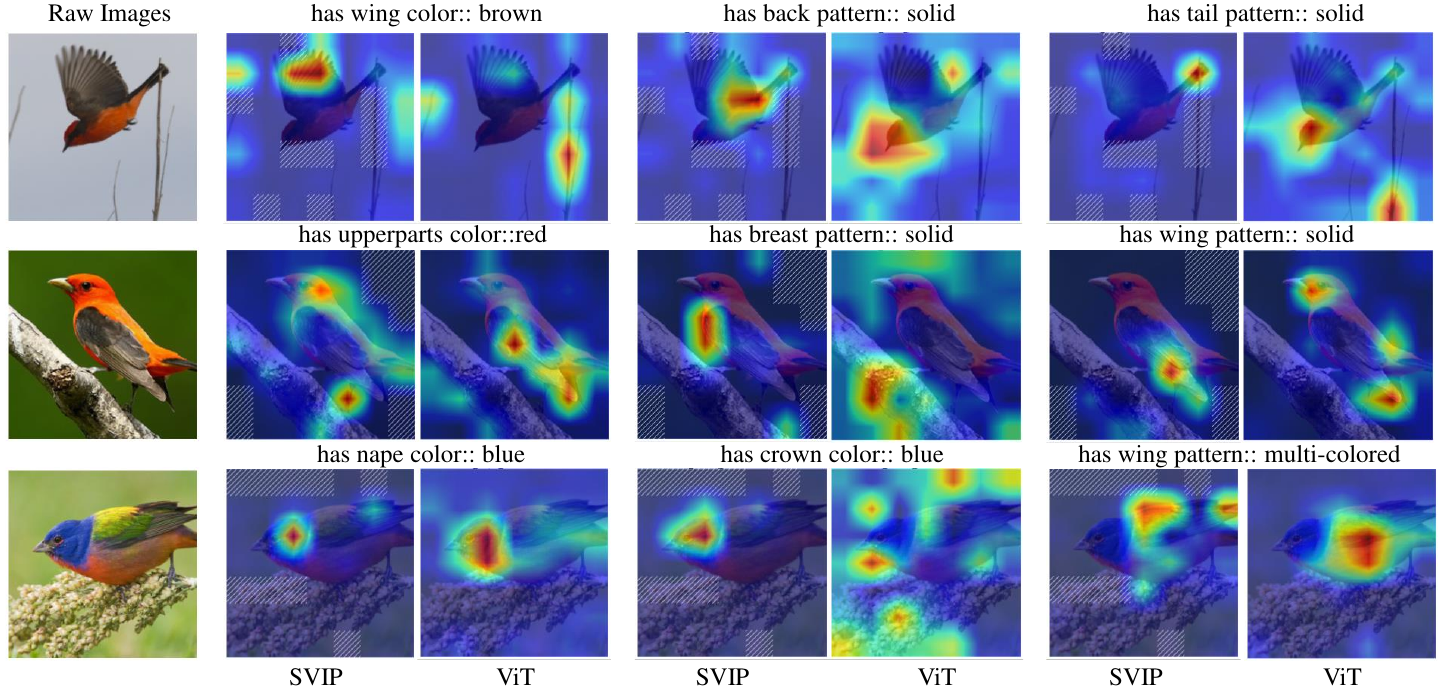}
    \vspace{-13pt}
    \caption{Visualization of the predicted attribute attention maps on three random test samples of CUB dataset.}
    \label{fig:att}
    \vspace{-10pt}
\end{figure*}

\subsection{Main Results}
\noindent \textbf{Effectiveness for Semantic Misalignment.} 
To show the effectiveness of SVIP in addressing semantic misalignment, we compare SVIP with methods aim to achieve this either in the feature space (RFF\cite{han2020learning}, SDGZSL\cite{chen2021semantics}, Dis‐VAE\cite{li2021generalized}) or at the model level (ZSLViT\cite{chen2024progressive}). As shown in Table \ref{gzslperoformance}, our method achieves superior performance under both the ZSL and GZSL settings on three benchmark datasets, except on ZSL performance where we achieve comparable performance on AwA2. These improvements confirm that filtering out irrelevant patches at the input stage is highly effective for strengthening visual–semantic alignment and boosting overall zero‐shot recognition accuracy.

\vspace{5pt}
\noindent \textbf{Comparison with CNN Backbones.} 
We next first compare SVIP with a group of methods that adopt ResNet101 as their backbone (upper block of Table \ref{gzslperoformance}). Specifically, approaches such as APN \cite{xu2020attribute}, GEM \cite{liu2021goal}, MSDN \cite{chen2022msdn},  Dis‐VAE, and ICCE all operate on global CNN‐extracted features to align visual and semantic spaces for zero‐shot learning. Despite their efficacy on certain datasets, our SVIP achieves consistently state-of-the-art performance results across CUB, AwA2, and SUN. Notably, SVIP obtains an H value of 75.0\% on CUB, 74.9\% on AwA2, and 50.7\% on SUN, surpassing all competing ResNet101‐based methods. These results demonstrate SVIP’s effectiveness in capturing fine‐grained semantic cues while maintaining strong discriminative ability for both seen (S) and unseen (U) classes.

\vspace{3pt}
\noindent \textbf{Comparison with ViT Backbones.}
We next compare SVIP with recent approaches with ViT as backbone, including language-driven methods (CLIP \cite{radford2021learning}, CoOp \cite{zhou2022learning}, I2DFormer \cite{naeem2022i2dformer}, I2MV \cite{naeem2023i2mvformer}), attribute-based methods (DUET\cite{chen2023duet}, ZSLViT \cite{chen2024progressive}). Generally, ViT‐based methods benefit from attention mechanisms over patch embeddings, but often lack explicit supervision for semantic‐unrelated patches. 
As shown, SVIP delivers the best performance in terms of unseen accuracy (U) on CUB (72.1\%) and SUN (53.7\%), as well as a leading harmonic mean on both datasets. On AwA2, our approach also shows competitive results, attesting to the adaptability of SVIP’s localized attribute reasoning across a range of visual domains.

\vspace{3pt}
\noindent \textbf{Ablation Study.}
We conduct an ablation study to analyze the contribution of each component in SVIP (Table \ref{Tab:ablations}). The Baseline (ViT) uses a simple linear layer on the class token without patch-level operations, yielding the lowest performance. SVIP w/o SSPS removes the Self-Supervised Patch Selection, applying semantic patches indiscriminately, leading to a significant drop, highlighting the necessity of targeted patch selection. SVIP w/o PSC directly removes semantic-unrelated patches instead of contextualizing them, which slightly disrupts object structure and lowers performance. SVIP w/o JSD excludes Jensen–Shannon divergence, reducing training stability and overall accuracy. SVIP w/o W2P replaces word embeddings with randomly initialized patch embeddings, resulting in degraded semantic alignment. SVIP w/o P2A removes Patch-to-Attribute mapping, forcing reliance on the class token, which weakens unseen class accuracy. 

\noindent \textbf{Hyper-parameter sensitivity.}
In a series of experiments, we investigate the influence of the four hyper-parameters on the performance of our proposed method. Figure \ref{hyper} (a) shows the effect of varying the number of preserved patches $M$ from 49 to 35.
The performance saturation occurs at around 40, and larger patch numbers tend to decrease performance due to over-pruned semantic-related patches.
Figure \ref{hyper} (b) demonstrates that the JS Divergence is a considerably sensitive hyperparameter, achieving the best performance at 1.
In Figure \ref{hyper} (c), we vary the loss coefficient for patch classfier $\lambda_2$ and found that both ZSL and GZSL achieve the best performance at 3. 
In Figure \ref{hyper} (d), we found that the temperature scale has a considerable impact on the performance variation, which peaks at 5.

\begin{figure}[t]
    \centering
    \includegraphics[width=1.0\linewidth]{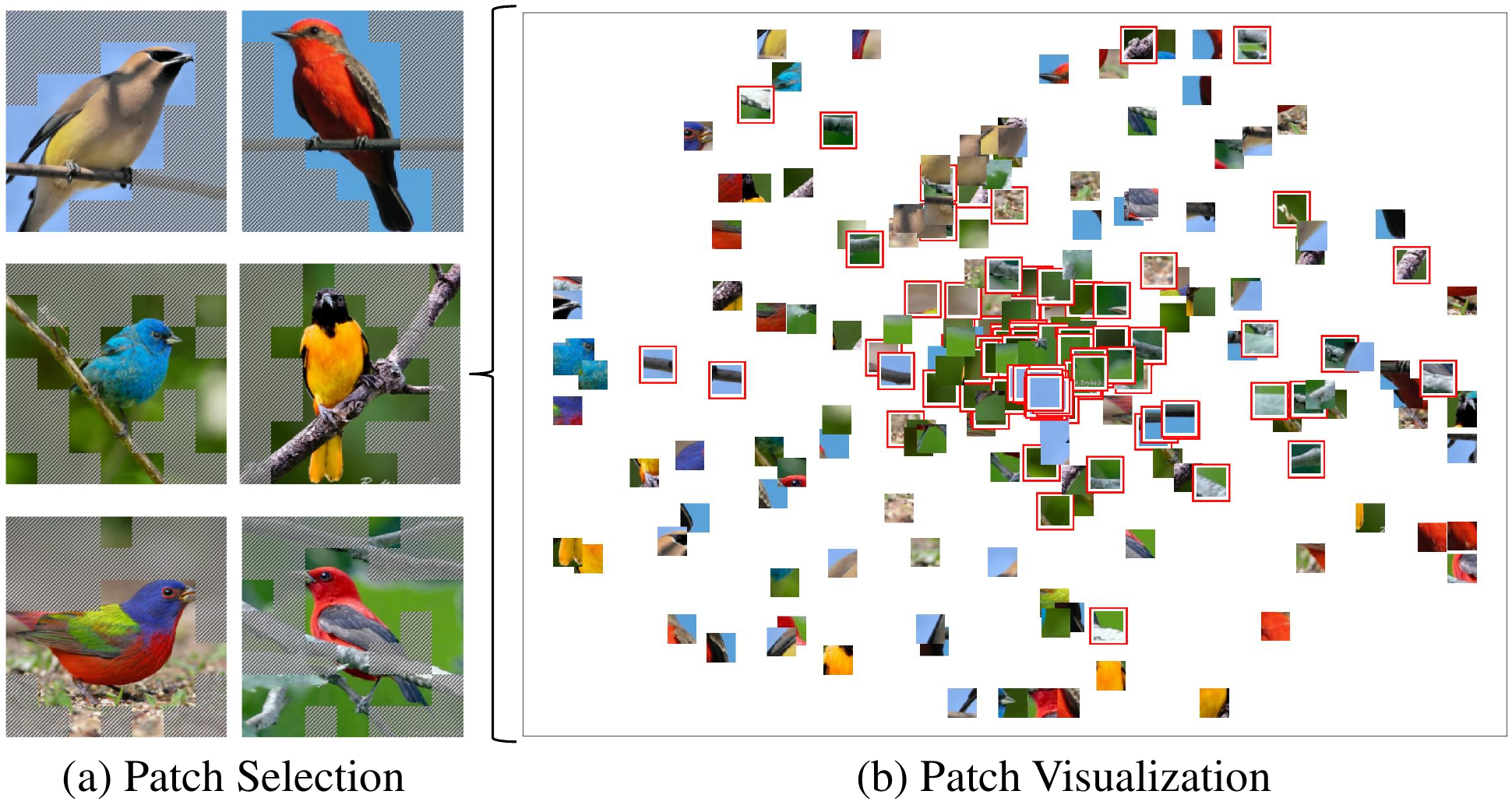}
    \caption{(a) Example images with semantic‐unrelated patches masked. (b) t-SNE visualization of the image patches, where red bounding boxes indicate the semantically unrelated patches.}
    \label{fig:vis_token}
    \vspace{-10pt}
\end{figure}

\subsection{Qualitative Study}
\noindent \textbf{Attribute Localization.}
We present attention maps (\textit{i.e.,} the layer before max pooling operation) in Figure \ref{fig:att}, for randomly selected attributes on three example images from the CUB dataset. The attention maps are generated by applying max pooling operations to the model output, highlighting the regions where the model is focusing on each attribute. The regions that are masked with striped hatch are identified as semantic-unrelated patches and thus not considered for attribute localization.
The baseline method ViT focuses on many incorrect patches that are classified as semantic-unrelated patches in SVIP, which demonstrates the usefulness of SSPS.

\begin{figure}[t]
    \centering
    \includegraphics[width=1.0\linewidth]{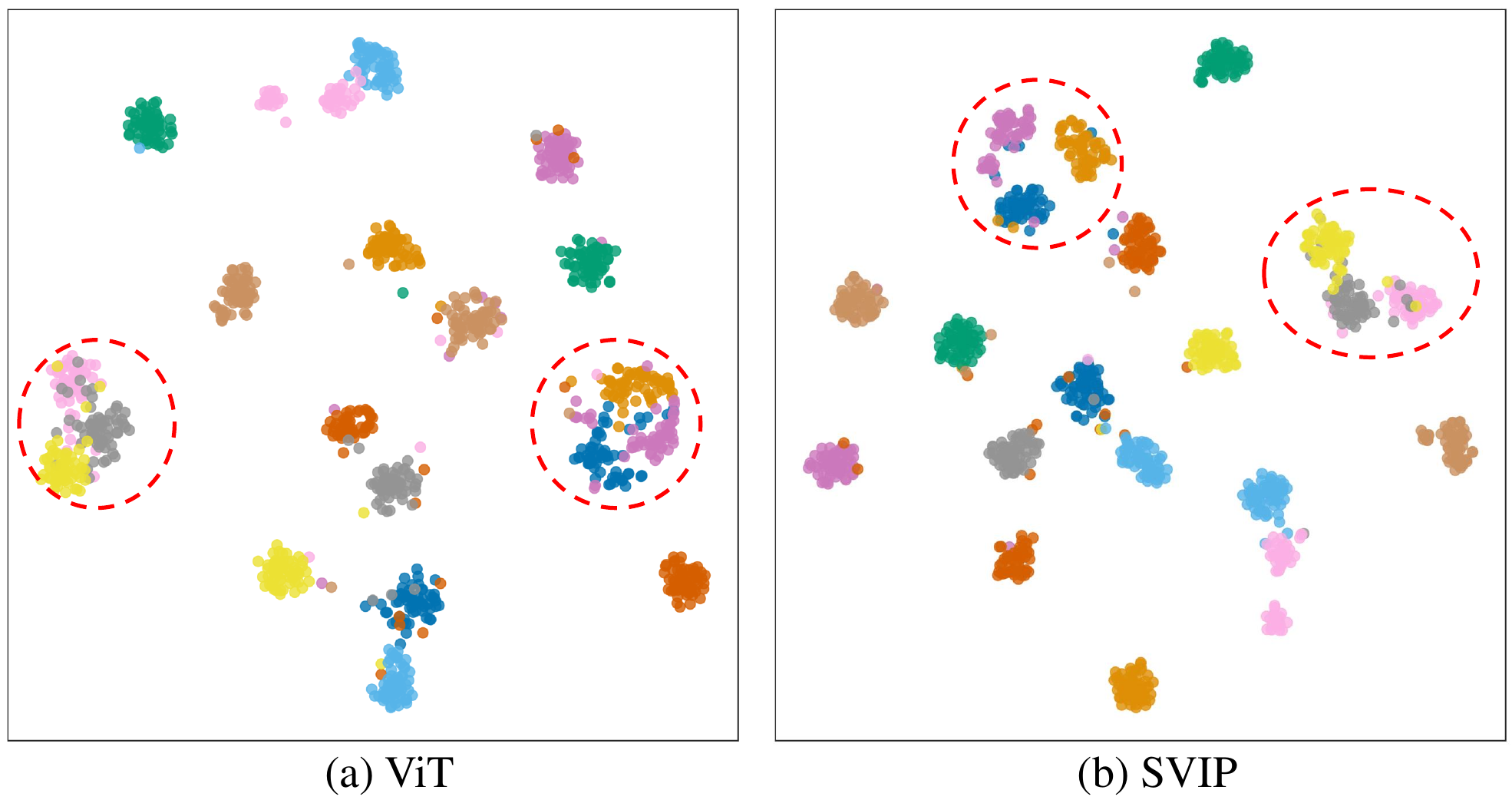}
    \caption{t-SNE visualization of the predicted attribute vectors from (a) baseline ViT and (b) our proposed SVIP. }
    \label{fig:tsne}
    \vspace{-15pt}
\end{figure}

\vspace{3pt}
\noindent \textbf{Attribute Prediction Visualization.}
Figure \ref{fig:tsne} presents a t-SNE visualization of the predicted attribute vectors from the baseline ViT (left) and our proposed SVIP (right) on images of 20 classes randomly picked from the unseen set of CUB dataset. In the ViT baseline Figure \ref{fig:tsne}(a), clusters appear less distinct, with overlapping and scattered attribute points, as indicated by the red dashed circles. This suggests that semantic-unrelated information disrupts the learned representations, leading to weaker visual-semantic alignment. In contrast, SVIP in Figure \ref{fig:tsne}(b) demonstrates more compact and well-separated clusters, where attributes are more clearly delineated. The red dashed circles highlight how SVIP reduces semantic ambiguity, ensuring that similar attributes form coherent and better-structured groups. These results validate the effectiveness of self-supervised patch selection and semantic contextualization in improving feature disentanglement, leading to stronger attribute-based reasoning in zero-shot learning.

\vspace{3pt}
\noindent \textbf{Patch Classifier Feature Visualization.}
Figure \ref{fig:vis_token} illustrates how our method selects and contextualizes semantic‐unrelated patches. In (a), each bird image is partitioned into patches, and those deemed irrelevant are shaded (diagonal stripes). This early‐stage filtering ensures that the model focuses on attributes crucial for bird identification (\textit{e.g.,} distinctive color patterns or shapes). In (b), we use t-SNE to visualize the embeddings in the intermediate layer in the patch classifier in a two‐dimensional space. We observe that semantic-unrelated patches cluster together as indicated in red bounding boxes, while patches capturing semantically relevant traits are more dispersed. This separation corroborates our patch classifier's ability to preserve meaningful visual cues and suppress extraneous information.

\section{Conclusions}
In this paper, we introduced a novel ViT-based framework for ZSL that enhances visual-semantic alignment by addressing the semantic misalignment issue in the input space. Our approach consists of two key components: self-supervised patch selection, which leverages aggregated attention scores to identify and suppress semantic-unrelated patches, and Semantic Contextualization, which replaces these patches with learnable attribute-aware embeddings. Extensive experiments on ZSL benchmarks demonstrate that our method achieves state-of-the-art performance results. Our work highlights the importance of early-stage semantic refinement in ZSL and suggest new directions for incorporating semantic priors into transformer-based vision models. Future work will explore adaptive strategies for patch selection and contextualization, allowing the model to dynamically adjust based on dataset characteristics. 

\section*{Acknowledgment}
This work is supported by the Australian Research Council under the streams of
Discovery Project (No. DP240101814).

{
\small
\bibliographystyle{ieeenat_fullname}
\bibliography{main}
}

\end{document}